\documentclass[sigconf, preprint, nonacm]{acmart}

\renewcommand\footnotetextcopyrightpermission[1]{} 
\usepackage{multirow}
\usepackage{amsmath}

\usepackage[table]{xcolor} 
\definecolor{tblgray}{gray}{0.9}

\usepackage{tcolorbox}

\tcbuselibrary{skins, breakable}

\usepackage{enumitem}

\usepackage{tabularx} % 用于自动计算列宽，防止表格出页
\usepackage{booktabs} % 用于三线表 (\toprule, \midrule, \bottomrule)

\AtBeginDocument{%
  \providecommand\BibTeX{{%
    \normalfont B\kern-0.5em{\scshape i\kern-0.25em b}\kern-0.8em\TeX}}}

\begin{document}

%%
%% 1. TITLE: MERIT
%%
\title{MERIT: Memory-Enhanced Retrieval for Interpretable Knowledge Tracing}

% \author{Firstname Lastname}
% \email{author1@email.com}
% \affiliation{%
%   \institution{Institution Name}
%   \city{City}
%   \country{Country}
% }

% \author{Firstname Lastname}
% \email{author2@email.com}
% \affiliation{%
%   \institution{Institution Name}
%   \city{City}
%   \country{Country}
% }

\author{Runze Li}
\authornote{Both authors contributed equally to this research.}
\affiliation{\institution{School of Computer Science and Technology, East China Normal University}\country{China}}
% \email{51265901021@stu.ecnu.edu.cn}

\author{Kedi Chen}
\authornotemark[1]
\affiliation{\institution{School of Computer Science and Technology, East China Normal University}\country{China}}
\affiliation{\institution{Shanghai Innovation Institute}\country{China}}
% \email{...}

\author{Guwei Feng}
\affiliation{\institution{School of Computer Science and Technology, East China Normal University}\country{China}}

\author{Mo Yu}
\affiliation{\institution{WeChat AI, Tencent}\country{China}}

\author{Jun Wang}
\authornote{Corresponding author.}
\affiliation{\institution{School of Computer Science and Technology, East China Normal University}\country{China}}

\author{Wei Zhang}
\authornotemark[2]
\affiliation{
  \institution{School of Computer Science and Technology, East China Normal University}
  \country{China}
}
\affiliation{
  \institution{Shanghai Innovation Institute}
  \country{China}
}

\begin{abstract}
Knowledge Tracing (KT) models students' evolving knowledge states to predict future performance, serving as a foundation for personalized education. While traditional deep learning models achieve high accuracy, they often lack interpretability. Large Language Models (LLMs) offer strong reasoning capabilities but struggle with limited context windows and hallucinations. Furthermore, existing LLM-based methods typically require expensive fine-tuning, limiting scalability and adaptability to new data. We propose \textbf{MERIT} (Memory-Enhanced Retrieval for Interpretable Knowledge Tracing), a training-free framework combining frozen LLM reasoning with structured pedagogical memory. Rather than updating parameters, MERIT transforms raw interaction logs into an interpretable memory bank. The framework uses semantic denoising to categorize students into latent cognitive schemas and constructs a paradigm bank where representative error patterns are analyzed offline to generate explicit Chain-of-Thought (CoT) rationales. During inference, a hierarchical routing mechanism retrieves relevant contexts, while a logic-augmented module applies semantic constraints to calibrate predictions. By grounding the LLM in interpretable memory, MERIT achieves state-of-the-art performance on real-world datasets without gradient updates. This approach reduces computational costs and supports dynamic knowledge updates, improving the accessibility and transparency of educational diagnosis.
\end{abstract}

%%
%% CCS Concepts & Keywords
%    <concept>
%        <concept_id>10010405.10010489.10010491</concept_id>
%        <concept_desc>Applied computing~Interactive learning environments</concept_desc>
%        <concept_significance>500</concept_significance>
%    </concept>
%    <concept>
%        <concept_id>10010147.10010178</concept_id>
%        <concept_desc>Computing methodologies~Artificial intelligence</concept_desc>
%        <concept_significance>500</concept_significance>
%    </concept>
%  </ccs2012>

% \ccsdesc[500]{Applied computing~Interactive learning environments}
% \ccsdesc[500]{Computing methodologies~Artificial intelligence}




\maketitle

%%
%% INTRODUCTION
%%
\section{Introduction}
\label{sec:intro}

Knowledge Tracing (KT) is a fundamental task in intelligent tutoring systems, designed to model students' evolving knowledge states and predict future performance \cite{Corbett2005KnowledgeTM}. Over the past decade, Deep Learning-based Knowledge Tracing (DLKT) has achieved high predictive accuracy by using Recurrent Neural Networks (RNNs) \cite{piech2015deep} and Transformer architectures \cite{ghosh2020context, shen2021learning} to capture complex temporal dependencies. Despite these successes, parametric models face two critical bottlenecks that limit their deployment in real-world educational settings. First, they operate as opaque systems, offering high-precision predictions but lacking the pedagogical rationale—such as identifying specific misconceptions—required for actionable intervention \cite{Minn2021InterpretableKT}. Second, they rely on extensive training over large datasets. This dependency incurs high computational costs and results in rigid models; adapting to new students or incorporating incremental data typically requires expensive retraining or fine-tuning \cite{song2022survey}, which risks catastrophic forgetting of previously learned patterns \cite{Im2023ForgettingawareLB}.

Large Language Models (LLMs) offer a potential solution through their robust reasoning capabilities and natural language understanding \cite{lee2024languagemodelknowledgetracing}. However, applying LLMs directly to KT introduces a "context-memory dilemma": generic pre-training lacks domain-specific educational nuances, while limited context windows prevent the processing of long-term student histories. Current solutions often resort to fine-tuning LLMs on educational data \cite{li2025ciktcollaborativeiterativeknowledge}, which ironically reintroduces the high computational costs and static knowledge issues that LLMs were intended to solve.

Retrieval-Augmented Generation (RAG) and memory-augmented networks have transformed how models handle knowledge-intensive tasks \cite{DBLP:conf/nips/LewisPPPKGKLYR020}. By separating reasoning (fluid intelligence) from knowledge storage (crystallized intelligence), these paradigms allow systems to access external information dynamically without parameter updates. This offers a promising yet underexplored direction for KT: if an LLM can retrieve relevant, interpretable cognitive patterns from external memory, it could theoretically perform precise diagnosis without training, while naturally handling incremental data by updating the memory bank.

Based on this insight, we propose MERIT (Memory-Enhanced Retrieval for Interpretable Knowledge Tracing), a training-free framework that moves KT closer to evidence-based diagnosis. Rather than fine-tuning an LLM to memorize student behaviors, MERIT uses a frozen LLM to reason over a structured, interpretative memory bank. This bank is constructed offline by converting raw interaction logs into "Annotated Cognitive Paradigms", namely explicit Chain-of-Thought (CoT) traces that explain why a student succeeded or failed. During inference, MERIT employs a hierarchical retrieval strategy to identify historical peers with similar cognitive schemas and injects their reasoned trajectories into the LLM's context. This "retrieve-then-reason" paradigm grounds predictions in pedagogical evidence and helps reduce mitigating hallucinations and momentum bias.

MERIT addresses the limitations of prior approaches by eliminating the need for gradient updates. It serves as a plug-and-play solution that reduces deployment costs and latency. Crucially, it resolves the incremental data challenge: new student interactions can be instantly added to the memory bank, allowing the system to adapt continuously without retraining. Furthermore, by retrieving human-readable reasoning traces rather than latent vectors, MERIT improves transparency and provides educators with interpretable diagnostic signals alongside performance predictions.

Our main contributions are summarized as follows:
\begin{itemize}[leftmargin=*]
    \item We propose a cost-effective, training-free framework that bypasses LLM fine-tuning. By utilizing non-parametric memory retrieval, MERIT achieves state-of-the-art performance using only inference-time compute, ensuring scalability.
    \item We introduce a memory construction pipeline that transforms raw logs into structured cognitive paradigms. This converts the black-box prediction task into an interpretable, evidence-based reasoning process.
    \item Our architecture supports incremental learning. The decoupling of reasoning and memory allows for seamless integration of new data, resolving the rigidity and maintenance costs of traditional DLKT models.
    \item Experiments on four diverse real-world datasets demonstrate that MERIT significantly outperforms both traditional deep learning baselines and standard LLM methods, showing that organized, interpretable memory is superior to parameter optimization in data-sparse educational contexts.
\end{itemize}

\section{Related Work}
\label{sec:related}

\subsection{Evolution of Knowledge Tracing and Interpretability Bottlenecks}
Knowledge tracing (KT) is a fundamental task in educational data mining that models students' evolving knowledge states to predict future performance \citep{Corbett2005KnowledgeTM}. Early approaches, such as Bayesian Knowledge Tracing (BKT) \citep{Corbett2005KnowledgeTM} and its extensions \citep{Pardos2011KTIDEMII}, used probabilistic graphical models to infer latent mastery. While other machine learning methods \citep{Pavlik2009PerformanceFA} also attempted to capture learning curves, they often struggled to model complex temporal dependencies and hidden concept interactions.

Deep Learning-based Knowledge Tracing (DLKT) emerged to address these limitations by using neural networks to extract latent representations from student interaction sequences \citep{piech2015deep,song2022survey,DBLP:conf/aaai/00060HL023}. Seminal works like DKT \citep{Piech2015DeepKT} introduced RNNs to the domain, while DKVMN \citep{Zhang2016DynamicKM} incorporated memory networks to enhance state tracking, establishing a foundation for subsequent research \citep{Liu2019EKTEK, Nagatani2019AugmentingKT}. The advent of the Transformer architecture further advanced the field. Models such as SAKT \citep{Pandey2019ASA}, LPKT \citep{Shen2021LearningPK}, and FoLiBi \citep{Im2023ForgettingawareLB} have significantly improved predictive accuracy by capturing long-range contextual dependencies and cognitive dynamics. Recently, Large Language Models (LLMs) have also been adapted for KT tasks \citep{cho2024systematicreviewknowledgetracing,lee2024languagemodelknowledgetracing}, using their natural language processing capabilities to handle complex learning scenarios.

Despite these advancements, transparency remains a critical challenge. While inherently explainable architectures \citep{Minn2021InterpretableKT} and post-hoc analysis methods \citep{cui2024interpretableknowledgetracingresponse} offer some insights, many DLKT models remain opaque, making auditing difficult in high-stakes educational contexts. Although LLMs can generate natural language rationales via chain-of-thought prompting \citep{xu2025largereasoningmodelssurvey,chen2025surveyinductivereasoninglarge} or self-explanation strategies \citep{DBLP:conf/acl/YugeswardeenooZO24,aswani2024autoevolveenhancinglargelanguage}, and internal probing tools like DeepDecipher \citep{garde2023deepdecipheraccessinginvestigatingneuron} reveal neuron-level behaviors \citep{paulo2025automaticallyinterpretingmillionsfeatures,duan-etal-2025-unveiling}, applying these techniques to KT is often costly. Existing explainable methods typically require extensive training or fine-tuning to align the model with KT tasks \citep{moghaddam2025explainableknowledgegraphretrievalaugmented,li2025ciktcollaborativeiterativeknowledge,lee2026training}, which consumes excessive computational resources and limits their generalizability.

\subsection{Retrieval-Augmented Generation and Memory Mechanisms}
Retrieval-Augmented Generation (RAG) and memory-augmented networks have gained prominence as solutions to the dependency on large-scale training data and the rigidity of parametric models \citep{ha2018memory,DBLP:conf/nips/LewisPPPKGKLYR020}. Unlike traditional deep learning models that rely solely on internal weights, these approaches allow systems to dynamically access and incorporate external knowledge bases. This paradigm reduces reliance on massive training datasets and facilitates the handling of incremental data, as the model can update its knowledge by simply refreshing the retrieval memory without retraining.

In educational contexts, memory-augmented methods offer a pathway to enhanced interpretability. By explicitly storing and revisiting student historical learning trajectories \citep{wang2023continuous,he2023man}, these models ground their predictions in retrieved evidence rather than abstract latent vectors. Recent work has demonstrated that memory modules can represent intermediate student states more intuitively \citep{ma2025knowledge}, helping educators understand learning progress. Building on these advantages, our work introduces a framework that is inherently interpretable and training-free. By rendering the LLM's reasoning transparent through retrieved cognitive patterns, we transform binary correctness predictions into deep diagnostics. Unlike prior methods that require heavy training, our approach serves as a plug-and-play solution, providing readable, evidence-based support for educational decision-making with minimal computational overhead.

%% 主框架图插入代码
%% 建议：将文件名 "kdd2026图.pdf" 重命名为 "framework.pdf" 以避免中文路径编译报错
\begin{figure*}[!t]
  \centering
  \includegraphics[width=0.95\textwidth]{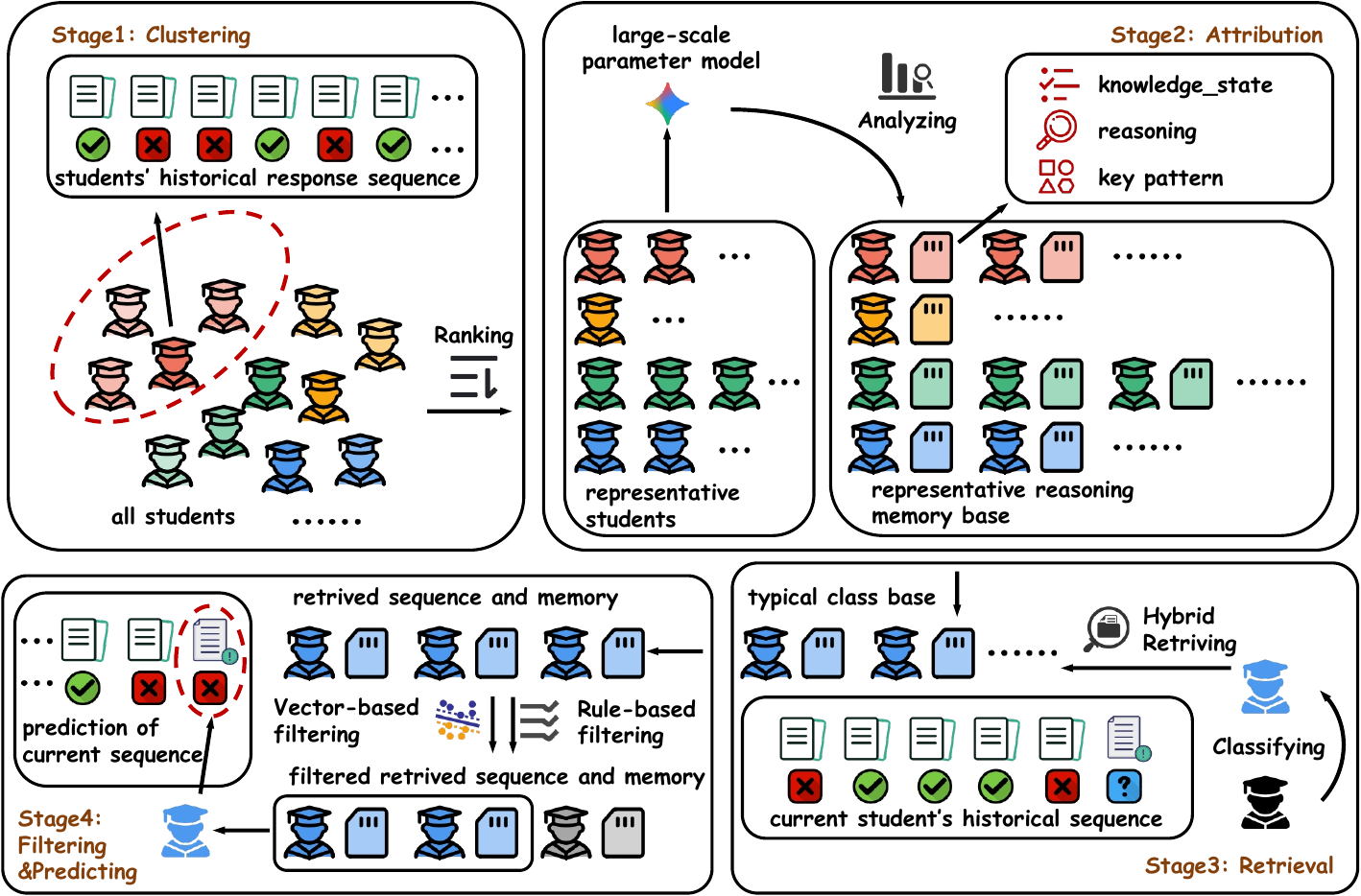}
  \caption{Overview of the MERIT framework. The architecture integrates a frozen LLM with an external cognitive memory across four stages. \textbf{Cognitive Schema Discovery} first discretizes the student space using semantic denoising and density-based clustering. \textbf{Interpretative Memory Bank Construction} then transforms expert reasoning traces into a static retrieval database. Subsequently, \textbf{Hierarchical Cognitive Retrieval} applies global centroid routing and local hybrid search to maintain domain consistency. Finally, \textbf{Logic-Augmented Reasoning} incorporates semantic difficulty calibration and explicit boundary constraints to regulate the prediction.}
  \label{fig:framework}
\end{figure*}

\section{Methodology}

\subsection{Framework Overview}
The MERIT framework addresses the challenges of interpretability and stability in Knowledge Tracing through a pipeline composed of four distinct stages, as illustrated in Figure \ref{fig:framework}. The process begins with \textit{Cognitive Schema Discovery}, where raw student interaction logs are projected into a semantic space and grouped into latent clusters to manage the diversity of learning behaviors. In the second stage, \textit{Attribution and Memory Construction}, we select representative samples from these clusters and employ a large model to generate explicit reasoning traces, including knowledge states and behavioral patterns. These annotated examples form a structured paradigm bank.

During the online phase, the system proceeds to the \textit{Retrieval} stage. For a target student, the model identifies the most relevant cognitive cluster and fetches similar historical paradigms using a combination of semantic vectors and keyword matching. Finally, the \textit{Filtering and Prediction} stage synthesizes these retrieved contexts. Here, the system applies vector-based and rule-based constraints—such as difficulty consistency checks—to filter potential hallucinations before the student model generates the final performance prediction. This decoupled design combines the precision of retrieval with the reasoning capabilities of large language models.

\subsection{Problem Formulation}
We define the Knowledge Tracing (KT) task as a sequential prediction problem. Let $\mathcal{U}$ denote the set of students and $\mathcal{E}$ the set of exercises. A student's learning trajectory is represented as a sequence $\mathcal{S}_u = \{s_1, s_2, \dots, s_t\}$, where each interaction step $s_k = (e_k, r_k, d_k)$ consists of the exercise text $e_k$, the binary response correctness $r_k \in \{0, 1\}$, and a continuous difficulty scalar $d_k \in [0, 1]$. The objective is to estimate the probability $\hat{y}_{t+1} = P(r_{t+1}=1 \mid e_{t+1}, d_{t+1}, \mathcal{S}_u)$, which reflects the likelihood that the student will correctly answer the next exercise. Unlike traditional deep learning approaches that require extensive parameter updates, our framework operates in a training-free paradigm, using a frozen Large Language Model augmented by a retrieved interpretative memory.

\subsection{Stage 1: Cognitive Schema Discovery}
\label{sec:module1}

This stage aims to construct robust representations of student interactions and discretize the continuous student space into interpretable latent groups, termed \textit{Cognitive Schemas}. Educational logs typically contain high levels of noise—such as minor fluctuations in difficulty scores (e.g., $0.12$ vs. $0.13$)—that offer limited semantic value and may distract embedding models. We address this through a pipeline comprising semantic denoising, high-dimensional projection, and hierarchical clustering.

\subsubsection{Semantic Denoising Mechanism}
Raw interaction logs often mix pedagogical content (e.g., "Linear Algebra") with statistical metadata (e.g., "score: 0.5"). To prevent the model from overfitting to superficial numerical patterns, we apply a token regularization strategy. Let $T(\mathcal{S}_u)$ denote the textual serialization of student $u$'s interaction history. We define a filtering function $\phi(\cdot)$ that retains only alphabetic tokens of length $\ge 2$. Formally, the denoised sequence $\tilde{T}_u$ is obtained by:
\begin{equation}
\begin{split}
    \tilde{T}_u &= \phi(T(\mathcal{S}_u)) = \{w \mid w \in \text{Tokenize}(T(\mathcal{S}_u)), \\
    &\quad \quad \text{is\_alpha}(w) \land \text{len}(w) \ge 2\}
\end{split}
\end{equation}
This operation masks numerical digits and special characters, ensuring the downstream embedding model focuses on semantic concepts (e.g., "Fraction", "Geometry") rather than statistical artifacts, thereby capturing the intrinsic knowledge structure.

\subsubsection{Manifold Projection and Clustering}
We encode the denoised sequences using a large-scale embedding model optimized for semantic representation, yielding a high-dimensional vector via aggregation of token embeddings $\mathbf{h}_u \in \mathbb{R}^{d}$ for each student $u$.

Given that direct clustering in high-dimensional space is hindered by the curse of dimensionality, we adopt a hierarchical approach within a neural topic modeling framework. We first project $\mathbf{h}_u$ into a lower-dimensional manifold using Uniform Manifold Approximation and Projection (UMAP), which preserves local neighborhood structures. Subsequently, we identify latent groups $\mathcal{C} = \{C_1, C_2, \dots, C_K\}$ using a density-based clustering algorithm. Unlike partition-based methods (e.g., $k$-means) that assume spherical clusters, this density-based approach adapts to clusters of varying shapes, making it well-suited for the irregular distribution of student capabilities.

\subsubsection{Schema Representation via c-TF-IDF}
Each identified cluster $C_k$ corresponds to a specific \textit{Cognitive Schema} (e.g., "Proficient in Algebra but weak in Geometry"). To interpret these schemas, we compute the class-based TF-IDF (c-TF-IDF) score for each word $w$ within cluster $C_k$:
\begin{equation}
    W_{w, C_k} = \text{tf}_{w, C_k} \cdot \log \left(1 + \frac{A}{f_w}\right)
\end{equation}
Here, $\text{tf}_{w, C_k}$ denotes the frequency of word $w$ in cluster $C_k$, $A$ represents the average number of words per cluster, and $f_w$ is the frequency of word $w$ across all clusters. The top-ranked keywords serve as semantic labels for the schema, guiding the retrieval process in subsequent stages.

\subsection{Stage 2: Interpretative Memory Bank Construction}
\label{sec:module2}

Stage 1 discretizes the student space into broad schemas, yet these representations lack explicit pedagogical reasoning. Standard retrieval-augmented methods often rely on raw historical logs, which require the online model to perform complex implicit reasoning under tight latency constraints. MERIT addresses this by pre-computing and storing expert-level reasoning traces in an offline \textit{Interpretative Memory Bank}.

\subsubsection{Representative Prototype Selection}
Since annotating the entire dataset is computationally prohibitive, we use the cluster structure from Stage 1 to implement a centroid-based sampling strategy. For each cognitive schema $C_k$, we select the top-$K$ representative student sequences based on their proximity to the cluster centroid. This approach ensures the memory bank $\mathcal{B}$ captures distinct behavioral prototypes while filtering out noise.

\subsubsection{Generative Pedagogical Attribution}
We use a large-scale language model to perform retrospective analysis on the sampled sequences. Unlike the online phase, this offline process accesses the ground truth outcome $y_{t+1}$, allowing the model to transform the raw interaction tuple $(\mathcal{S}_{u}, y_{t+1})$ into a structured annotation $\mathcal{A}$.

Formally, for each selected sample, we generate a \textit{Memory Entry} $\mathcal{M} = \langle \mathcal{S}_{u}, y_{t+1}, \mathcal{A} \rangle$. The annotation $\mathcal{A}$ comprises four explanatory components. First, the \textit{Knowledge State} ($\mathbf{k}$) summarizes the student's mastery across concepts. Second, the \textit{Key Pattern} ($\mathbf{p}$) classifies behavior into archetypes from a label set $\mathcal{L}$:
\begin{equation}
\resizebox{0.85\columnwidth}{!}{%
$
\mathcal{L} = \{ \text{``Solid Mastery''}, \text{``Diff. Spike Failure''}, \text{``Careless Slip''}, \dots \}
$
}
\end{equation}
Third, the \textit{Difficulty Context} ($\mathbf{d}$) assesses the question's difficulty relative to the student's history. Finally, \textit{Causal Reasoning} ($\mathbf{r}$) provides a logical chain explaining the success or failure.

This transformation converts raw data points into "crystallized" pedagogical insights. The resulting Memory Bank $\mathcal{B} = \{ \mathcal{M}_1, \dots, \mathcal{M}_N \}$ functions as an external cognitive store, enabling the online model to access high-level reasoning patterns efficiently.

\subsection{Stage 3: Hierarchical Cognitive Retrieval}
\label{sec:module3}

Standard retrieval-augmented generation systems typically perform flat searches over monolithic databases. This approach often retrieves behaviorally irrelevant samples—such as mismatched error types—and increases computational latency. To mitigate this, we implement a hierarchical mechanism that integrates global routing with local hybrid search.

\subsubsection{Global Cognitive Routing via Centroid Matching}
The first phase restricts the search space to a domain-consistent subset. Using the cognitive schemas identified in Stage 1, we apply a divide-and-conquer strategy. For a target student $u$ with interaction sequence $\mathcal{S}_u$, we generate a query embedding $\mathbf{h}_u$ via the frozen encoder. We then assign the student to the optimal cognitive schema $C_{\hat{k}}$ by identifying the nearest cluster centroid:
\begin{equation}
    \hat{k} = \operatorname*{arg\,min}_{k \in \{1, \dots, K\}} \text{Dist}(\mathbf{h}_u, \mathbf{\mu}_k)
\end{equation}
where $\mathbf{\mu}_k$ represents the pre-computed centroid of cluster $k$. To address "cold-start" scenarios where a student's sequence is too short for reliable cluster assignment, the system routes queries to a global generic pool to ensure robustness.

\subsubsection{Partitioned Indexing and Hybrid Retrieval}
We partition the offline memory bank $\mathcal{B}$ into $K$ independent sub-indices $\mathcal{B}_1, \dots, \mathcal{B}_K$, reducing retrieval complexity from $O(N)$ to approximately $O(N/K)$. Within the target partition $\mathcal{B}_{\hat{k}}$, we execute a parallel hybrid search to retrieve the top-$N$ memory entries. This process combines dense and sparse signals to capture both semantic trends and specific concept overlaps.

We use the FAISS library to perform an inner-product search between the query vector $\mathbf{h}_u$ and stored memory vectors, capturing latent behavioral similarities. Complementing this, we apply the BM25Okapi algorithm to calculate lexical overlap between interaction texts, ensuring that retrieved examples involve similar mathematical concepts. The final relevance score $S(u, v)$ for a candidate entry $v$ is computed as:
\begin{equation}
    S(u, v) = \alpha \cdot \text{CosSim}(\mathbf{h}_u, \mathbf{h}_v) + (1-\alpha) \cdot \text{BM25}(\mathcal{S}_u, \mathcal{S}_v)
\end{equation}
We set the balancing factor $\alpha = 0.7$ and normalize both scores to the $[0, 1]$ range. The system returns the set $\mathcal{R}_u$ containing the $N$ entries with the highest combined scores, serving as the context for the subsequent inference stage.

\subsection{Stage 4: Logic-Augmented Reasoning and Prediction}
\label{sec:module4}

The final stage synthesizes the student's history with retrieved cognitive contexts to predict future performance. A primary challenge in applying LLMs to Knowledge Tracing is "Momentum Bias," where models tend to overgeneralize from recent sequences of correct answers on easy items, ignoring significant increases in difficulty. We address this through a framework that integrates semantic difficulty calibration, context filtering, and explicit boundary constraints.

\subsubsection{Semantic Difficulty Calibration}
Language models often struggle to interpret the semantic magnitude of raw floating-point values (e.g., distinguishing the conceptual gap between $0.65$ and $0.75$). We address this by mapping numerical difficulty scalars to explicit linguistic tags using a discrete function $L(\cdot)$. Let $d_t$ denote the difficulty at step $t$:
\begin{equation}
    L(d_t) = 
    \begin{cases} 
    \texttt{[EASY]} & \text{if } d_t < 0.35 \\
    \texttt{[MEDIUM]} & \text{if } 0.35 \le d_t < 0.70 \\
    \texttt{[HARD]} & \text{if } d_t \ge 0.70 
    \end{cases}
\end{equation}
Embedding these tokens directly into the input sequence guides the model to process difficulty as a distinct semantic feature rather than a statistical artifact, effectively calibrating its reasoning against the inherent complexity of the question.

\subsubsection{Contextual Quality Control}
Direct concatenation of retrieved results often introduces noise, as vector similarity does not guarantee pedagogical relevance. We therefore implement a post-retrieval filtering protocol to prevent irrelevant contexts from biasing the inference. For each candidate $\mathcal{M}_v \in \mathcal{R}_u$, we calculate a relevance confidence score $S(u, v)$. We discard candidates where this score falls below a similarity threshold $\tau$ or where the sequence length ratio $|\mathcal{S}_v|/|\mathcal{S}_u|$ differs substantially from 1. This mechanism ensures that the final prompt consists only of examples that share both semantic content and structural complexity with the target query, minimizing the risk of false attribution.

\subsubsection{Logic-Constrained Inference}
To regulate the generation process, we impose a hard logic constraint, termed the \textit{Spike Rule}, within the system prompt. This rule prevents over-optimistic extrapolation during sudden difficulty transitions. Formally, given a recent sequence of correct responses $r_{t-k:t}$ and a subsequent item difficulty $d_{t+1}$, the constraint is defined as:
\begin{equation}
    \text{If } (\forall i \in [t-k, t], r_i = 1) \land (L(d_{t+1}) = \texttt{[HARD]}) \implies P(\hat{y}_{t+1}=1) < \delta
\end{equation}
Here, $\delta$ is a dynamic threshold derived from the filtered memory bank. This constraint posits that success on simpler items does not imply mastery of significantly harder concepts. The final prediction is generated by the inference model $f_{\theta}$, which takes the calibrated history $\tilde{\mathcal{S}}_u$, the filtered context set $\mathcal{R}'_u$, and the logic constraints as input:
\begin{equation}
    \hat{y}_{t+1} = f_{\theta}(\tilde{\mathcal{S}}_u, \mathcal{R}'_u, d_{t+1})
\end{equation}
By grounding predictions in high-quality evidence and rigorous logic, MERIT balances the generalization capabilities of LLMs with precise, rule-based constraints.

\section{Experiments}

\subsection{Experimental Setup}

\subsubsection{Datasets and Preprocessing}
We evaluate MERIT on four widely used benchmarks—ASSISTments 2009\cite{feng2009addressing}, ASSISTments 2012\cite{feng2009addressing}, Eedi\cite{wang2020instructions}, and BePKT\cite{zhu2022programming}—covering both mathematical and programming domains to test robustness across diverse cognitive tasks. ASSISTments 2009 and ASSISTments 2012 originate from the ASSISTments platform; ASSISTments 2009 is characterized by sparsity and high concept overlap, whereas ASSISTments 2012 offers a denser interaction log with distinct problem styles. Eedi is a large-scale diagnostic dataset from the NeurIPS 2020 Education Challenge, featuring long interaction sequences that challenge the model's ability to capture long-term dependencies. BePKT is collected from an online programming judge platform. Unlike the math-focused datasets above, BePKT targets programming knowledge tracing, involving complex cognitive demands specific to coding. This dataset serves as a critical testbed for evaluating the generalizability of our approach beyond mathematics education.

To ensure data quality, we applied a standardized preprocessing pipeline. We filtered out students with interaction sequences shorter than 5 steps to eliminate insufficient data points. Sequences were truncated to a maximum length of 50 to accommodate LLM context window constraints while retaining recent behavioral trends. For evaluation, we adopted a strictly temporal 80/20 train-test split to prevent data leakage, ensuring a realistic assessment of predictive capability. Table \ref{tab:dataset_statistics} summarizes the statistics of the preprocessed datasets.

\begin{table}[!t]
\centering
\setlength{\tabcolsep}{4pt}
\resizebox{0.95\columnwidth}{!}{%
\begin{tabular}{lcccc}
\toprule
\textbf{Dataset} & \textbf{ASSIST2009} & \textbf{ASSIST2012} & \textbf{Eedi} & \textbf{BePKT} \\
\midrule
\textbf{\# Responses} & 0.4m & 2.7m & 17.8m & 23.9k \\
\textbf{\# Sequences} & 8.3k & 67.1k & 475.4k & 1.9k \\
\textbf{\# Questions} & 6.9k & 53.1k & 2.7k & 0.5k \\
\textbf{\# Concepts} & 200 & 265 & 386 & 100 \\
\bottomrule
\end{tabular}%
}
\caption{Statistics of the preprocessed datasets.}
\label{tab:dataset_statistics}
\end{table}

\subsubsection{Baselines}
We compare MERIT against nine baselines, categorized into mainstream Deep Learning Knowledge Tracing (DLKT) methods and LLM-enhanced frameworks.

\begin{table*}[t]
\centering
\caption{Overall performance comparison on ASSIST09, ASSIST12, Eedi, and BePKT. The best results are \textbf{bolded}, and the second best are \underline{underlined}. Methods marked with * are training-free.}
\label{tab:main_results}
\resizebox{\textwidth}{!}{
\begin{tabular}{l|c|ccc|ccc|ccc|ccc}
\toprule
\multirow{2}{*}{\textbf{Method}} & \multirow{2}{*}{\textbf{Category}} & \multicolumn{3}{c|}{\textbf{ASSIST09}} & \multicolumn{3}{c|}{\textbf{ASSIST12}} & \multicolumn{3}{c|}{\textbf{Eedi}} & \multicolumn{3}{c}{\textbf{BePKT}} \\
 & & AUC & ACC & F1 & AUC & ACC & F1 & AUC & ACC & F1 & AUC & ACC & F1 \\
\midrule
% --- DL-based Methods
DKT & DL-based & 0.7485 & 0.7315 & 0.7985 & 0.7512 & 0.7385 & 0.7993 & 0.7286 & 0.7065 & 0.8061 & 0.6985 & 0.6852 & 0.5647 \\
AKT & DL-based & 0.7684 & 0.7185 & 0.7925 & 0.7657 & 0.7374 & 0.7925 & 0.7388 & 0.7132 & 0.7994 & 0.6975 & 0.6845 & 0.5591 \\
SAKT & DL-based & 0.7582 & 0.7365 & 0.8005 & 0.7616 & 0.7361 & 0.8027 & 0.7411 & 0.7165 & 0.8082 & 0.7042 & 0.6904 & 0.5712 \\
LPKT & DL-based & 0.7533 & 0.7317 & 0.8038 & 0.7485 & 0.7248 & 0.8044 & 0.7457 & 0.7222 & 0.8106 & 0.6958 & 0.6807 & 0.5730 \\
IKT & DL-based & 0.7439 & 0.7194 & 0.7937 & 0.7583 & 0.7386 & 0.7795 & 0.7338 & 0.7135 & 0.7859 & 0.6885 & 0.6730 & 0.5614 \\
DIMKT & DL-based & 0.7635 & 0.7422 & 0.8006 & 0.7619 & 0.7405 & 0.7888 & 0.7484 & 0.7273 & 0.8115 & 0.7186 & 0.7035 & 0.5693 \\
DKVMN & DL-based & 0.7516 & 0.7173 & 0.7845 & 0.7551 & 0.7288 & 0.7524 & 0.7382 & 0.7139 & 0.7583 & 0.6978 & 0.6844 & 0.5726 \\
\midrule
% --- LLM-based Methods---
EPLF* & LLM-based & 0.8127 & 0.7013 & 0.7160 & 0.6964 & 0.7030 & 0.7582 & 0.6784 & 0.6970 & 0.7340 & 0.7235 & 0.6743 & 0.5743 \\
EFKT & LLM-based & 0.6110 & 0.6485 & 0.7359 & 0.6679 & 0.6303 & 0.6683 & - & - & - & - & - & - \\
2T-KT & LLM-based & 0.8132 & 0.7455 & \textbf{0.8065} & 0.7160 & 0.7280 & 0.7545 & - & - & - & - & - & - \\
\midrule
% --- MERIT (Modified with Subscripts) ---
\textbf{MERIT}$_{\textnormal{Gemini-2.5-Flash}}$* & Ours & \textbf{0.8244} & \underline{0.7554} & \underline{0.8054} & \underline{0.7778} & \textbf{0.7441} & \textbf{0.8122} & \underline{0.7969} & \underline{0.7603} & \underline{0.8139} & \underline{0.7827} & \textbf{0.7850} & \textbf{0.6228} \\
\textbf{MERIT}$_{\textnormal{GPT-4o}}$* & Ours & \underline{0.8188} & \textbf{0.7615} & 0.7941 & \textbf{0.7792} & \underline{0.7408} & \underline{0.8094} & \textbf{0.7996} & \textbf{0.7647} & \textbf{0.8167} & \textbf{0.8036} & \underline{0.7825} & \underline{0.5991} \\
\bottomrule
\end{tabular}
}
\end{table*}

\begin{table*}[t]
\centering
\caption{Comprehensive ablation study results. The \colorbox{tblgray}{shaded} rows denote the full MERIT framework, distinguishing it from the baseline (w/o Retrieval) and ablated variants.}
\label{tab:ablation_full}
% 定义一个稍微深一点的灰色以确保可见
\definecolor{tblgray}{gray}{0.9}
\resizebox{\textwidth}{!}{
\begin{tabular}{l|ccc|ccc|ccc|ccc}
\toprule
\multirow{2}{*}{\textbf{Method}} & \multicolumn{3}{c|}{\textbf{ASSIST09}} & \multicolumn{3}{c|}{\textbf{ASSIST12}} & \multicolumn{3}{c|}{\textbf{Eedi}} & \multicolumn{3}{c}{\textbf{BePKT}} \\
 & AUC & ACC & F1 & AUC & ACC & F1 & AUC & ACC & F1 & AUC & ACC & F1 \\
\midrule
% ================= Block 1: Gemini =================
Base (w/o Retrieval) & 0.6708 & 0.4794 & 0.3131 & 0.6749 & 0.4383 & 0.3601 & 0.7333 & 0.6855 & 0.7458 & 0.7392 & 0.6875 & 0.2038 \\
\hspace{1em}\textit{w/o Global Routing} & 0.6715 & 0.4782 & 0.3297 & 0.6804 & 0.4500 & 0.3838 & 0.7312 & 0.6815 & 0.7433 & 0.7461 & 0.6925 & 0.2166 \\
\hspace{1em}\textit{w/o Interpretative Traces} & 0.6709 & 0.4879 & 0.3482 & 0.6792 & 0.4487 & 0.3822 & 0.7295 & 0.6850 & 0.7460 & 0.7573 & 0.6800 & 0.1688 \\
\hspace{1em}\textit{w/o Logic Constraints} & 0.7015 & 0.5012 & 0.3622 & 0.6887 & 0.4613 & 0.4035 & 0.7329 & 0.6935 & 0.7675 & 0.7617 & 0.6950 & 0.2278 \\
\rowcolor{tblgray} % 灰色背景
\textbf{MERIT}$_{\textnormal{Gemini-2.5-Flash}}$ & \textbf{0.8244} & \textbf{0.7554} & \textbf{0.8054} & \textbf{0.7778} & \textbf{0.7441} & \textbf{0.8122} & \textbf{0.7969} & \textbf{0.7603} & \textbf{0.8139} & \textbf{0.7827} & \textbf{0.7850} & \textbf{0.6228} \\
\midrule
% ================= Block 2: GPT-4o =================
Base (w/o Retrieval) & 0.6079 & 0.6211 & 0.5940 & 0.6539 & 0.5957 & 0.6263 & 0.7423 & 0.6965 & 0.7517 & 0.6191 & 0.6950 & 0.2375 \\
\hspace{1em}\textit{w/o Global Routing} & 0.6312 & 0.6344 & 0.6234 & 0.6613 & 0.6041 & 0.6377 & 0.7387 & 0.6962 & 0.7507 & 0.6340 & 0.7000 & 0.2593 \\
\hspace{1em}\textit{w/o Interpretative Traces} & 0.5983 & 0.6138 & 0.5863 & 0.6511 & 0.5805 & 0.6001 & 0.7412 & 0.6970 & 0.7520 & 0.6220 & 0.7125 & 0.2945 \\
\hspace{1em}\textit{w/o Logic Constraints} & 0.7160 & 0.6743 & 0.6675 & 0.6630 & 0.6091 & 0.6454 & 0.7312 & 0.6933 & 0.7695 & 0.6203 & 0.7125 & 0.3114 \\
\rowcolor{tblgray} % 灰色背景
\textbf{MERIT}$_{\textnormal{GPT-4o}}$ & \textbf{0.8188} & \textbf{0.7615} & \textbf{0.7941} & \textbf{0.7792} & \textbf{0.7408} & \textbf{0.8094} & \textbf{0.7996} & \textbf{0.7647} & \textbf{0.8167} & \textbf{0.8036} & \textbf{0.7825} & \textbf{0.5991} \\
\bottomrule
\end{tabular}
}
\end{table*}

\paragraph{DLKT Methods:} This category includes pioneering approaches such as DKT \cite{piech2015deep} and the memory-augmented DKVMN \cite{zhang2017dynamic}. We also include Transformer-based architectures like SAKT \cite{pandey2019self}, which uses self-attention for interaction modeling, and AKT \cite{ghosh2020context}, which introduces context-aware attention with monotonic decay. Furthermore, we benchmark against LPKT \cite{shen2021learning}, which explicitly models learning and forgetting dynamics; IKT \cite{minn2022interpretable}, focusing on interaction-centric modeling; and DIMKT \cite{shen2022assessing}, which integrates question difficulty into the tracing process.

\paragraph{LLM-based and Pre-training Methods:} To evaluate the effectiveness of our retrieve-then-reason paradigm, we compare against three advanced frameworks using large language models or pre-training techniques. Thinking-KT \cite{lee2026training} is a recently proposed training-free framework that leverages large reasoning models with test-time scaling for unified prediction. EFKT \cite{li2025explainable}, or Explainable Few-shot Knowledge Tracing, uses the generative capabilities of LLMs to infer student states from limited history. Finally, 2T-KT \cite{li2025llm} employs a topic-aware architecture to capture the hierarchical structure of knowledge concepts. Comparing against these methods allows us to isolate the specific contributions of MERIT's interpretative memory and logic-augmented reasoning.

\subsubsection{Implementation Details}
MERIT uses a decoupled architecture. In the memory construction phase (Stage 2), we use Gemini-2.5-Pro \cite{comanici2025gemini} via API to generate pedagogical traces, sampling top $K_{bank}=100$ representative sequences per cognitive cluster to build the Interpretative Memory Bank. We restrict both cognitive schema discovery (Stage 1) and memory bank construction to the training set to prevent information leakage. For unseen students in the test set, we assign cognitive schemas based on the proximity of their embedding vectors to the cluster centroids established during training.
For online inference (Stage 3), we evaluate two backbones: Gemini-2.5-Flash \cite{comanici2025gemini} for efficiency and GPT-4o \cite{hurst2024gpt} for reasoning capability. We use Qwen3 Embedding 8B \cite{zhang2025qwen3} ($D=4096$) for semantic vectorization and index vectors using FAISS for efficient similarity search. During inference, we retrieve the top $N=3$ entries using a hybrid search weight of $\alpha=0.7$. As all model interactions occur via API, the framework supports lightweight deployment without requiring local high-performance GPUs.

\subsubsection{Evaluation Metrics}
We evaluate performance using three commonly used metrics. Area Under the ROC Curve (AUC) measures the model’s ability to rank positive instances above negative ones and is less sensitive to class imbalance. We also report Accuracy (ACC), computed using a decision threshold of 0.5, and the F1 score (F1), which captures the trade-off between precision and recall.

\begin{figure*}[t]
\centering
\begin{minipage}[b]{0.4\linewidth}
    \centering
    \includegraphics[width=\linewidth]{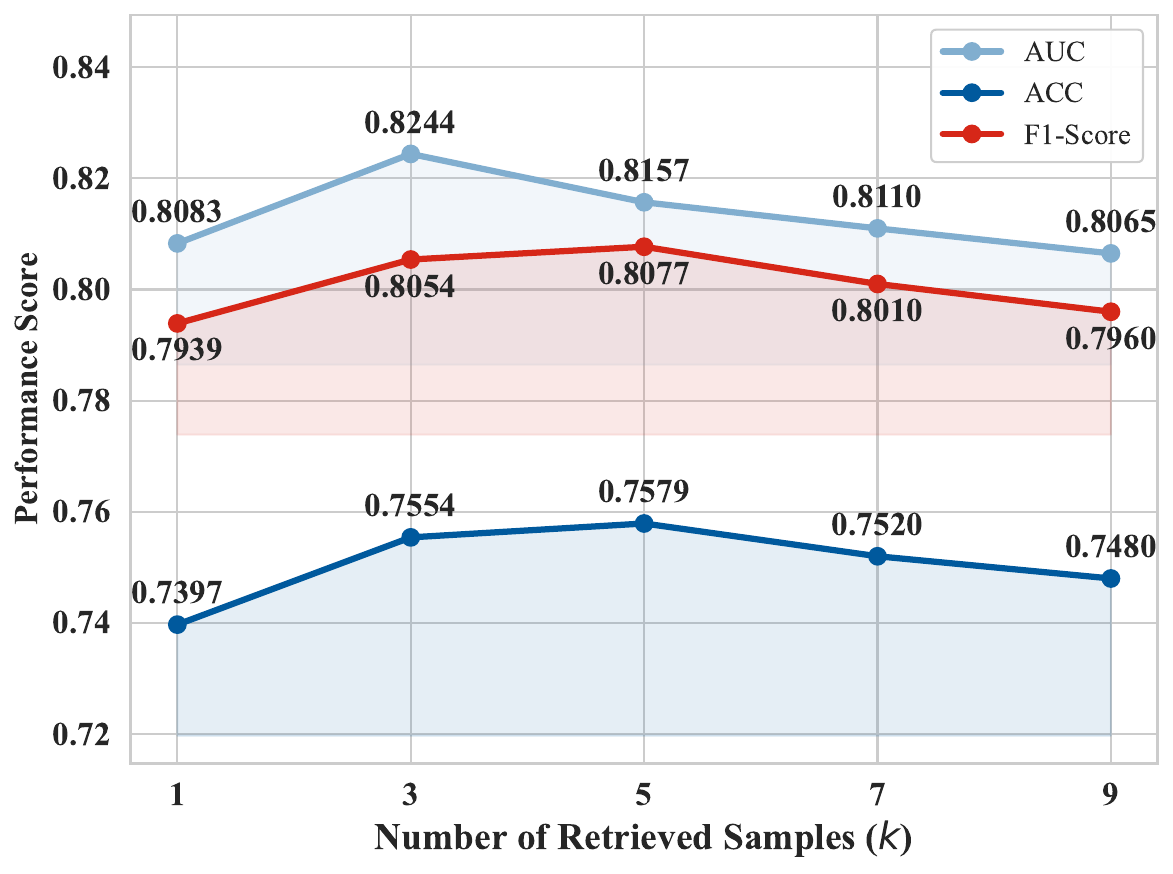}
    %\subcaption{Impact of retrieval volume $k$} % 如果用 subcaption 可以加子标题
\end{minipage}
\hspace{0.05\linewidth} % 两张图间距
\begin{minipage}[b]{0.4\linewidth}
    \centering
    \includegraphics[width=\linewidth]{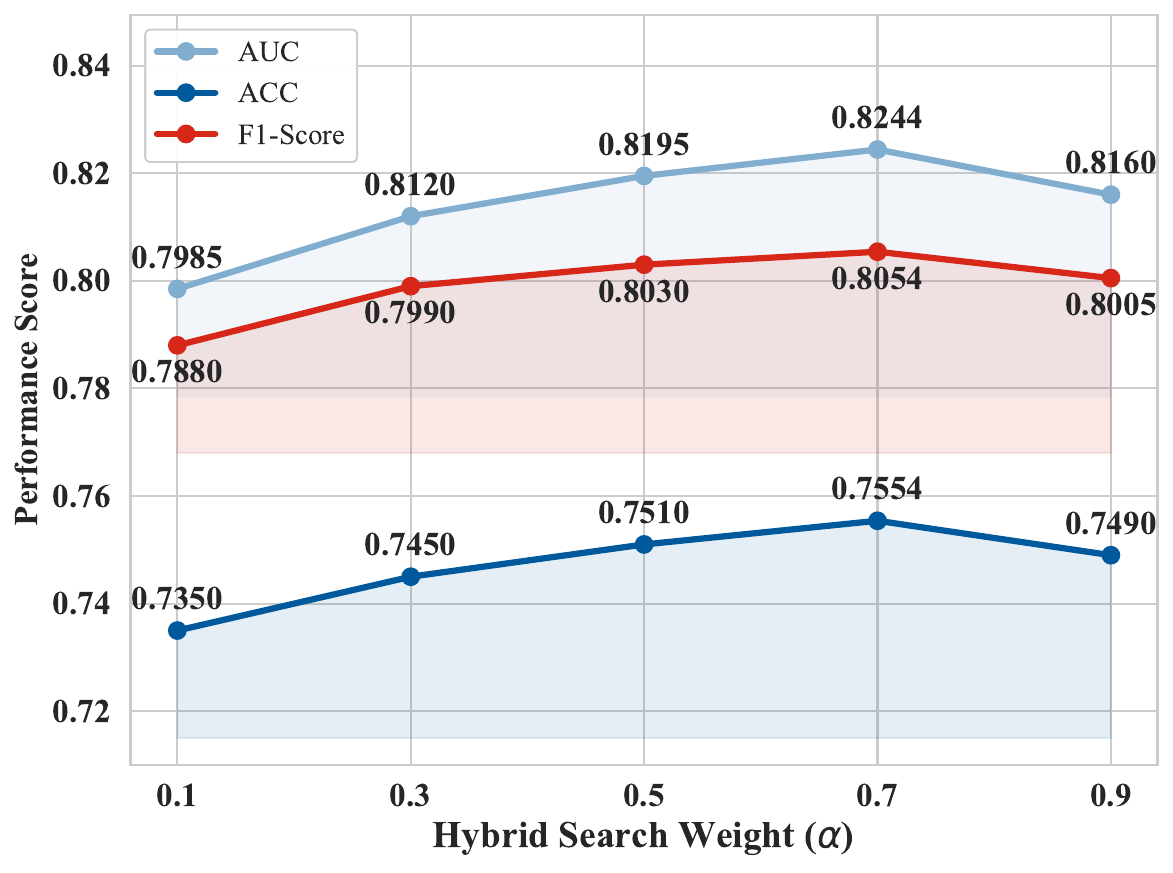}
    %\subcaption{Impact of hybrid search weight $\alpha$}
\end{minipage}

\caption{Parameter sensitivity analysis on ASSISTments 2009 using Gemini-2.5-Flash. (Left) Impact of retrieval volume $k$. Performance peaks at $k=3$, suggesting that a compact context window limits noise from irrelevant samples. (Right) Sensitivity to hybrid search weight $\alpha$. A weight of $0.7$ optimally balances latent semantic features with symbolic keyword matching.}
\label{fig:sensitivity}
\end{figure*}

\subsection{Main Results}

Table \ref{tab:main_results} presents the comparative results across all evaluated datasets, where MERIT consistently outperforms both traditional deep learning models and recent LLM-based frameworks.

\subsubsection{Performance Superiority}
The results in Table \ref{tab:main_results} indicate that MERIT establishes a new performance standard across the four benchmarks. On the ASSISTments 2009 dataset, the Gemini-backed variant achieves an AUC of 0.8244, which significantly surpasses the best deep learning baseline, AKT, at 0.7684. It also exceeds the strongest LLM-based competitor, 2T-KT, which scores 0.8132. This advantage remains consistent on the denser ASSISTments 2012 dataset. While traditional LLM approaches such as EPLF and 2T-KT suffer from generalization issues, leading to AUCs of 0.6964 and 0.7160, respectively, MERIT maintains robust performance with an AUC of 0.7778. These findings suggest that the retrieve-then-reason paradigm effectively integrates latent feature modeling with linguistic reasoning to ensure stability across varying data densities.

\subsubsection{Cross-Domain Generalization}
Performance on the BePKT dataset highlights the framework's adaptability to non-mathematical domains. Deep learning models typically plateau around an AUC of 0.70, exemplified by SAKT at 0.7042. In contrast, MERIT achieves an AUC of 0.8036 when using GPT-4o. This substantial improvement indicates that the semantic logic captured by the interpretative memory bank extends beyond algebraic rules to include the logical structures inherent in programming tasks. Standard sequence models often struggle to capture these complex cognitive dependencies without explicit reasoning signals.

\subsubsection{Backbone Robustness}
A comparison between the two backbone variants reveals that MERIT remains robust regardless of the inference engine. Although GPT-4o yields slightly higher metrics on Eedi and BePKT, the cost-effective Gemini-2.5-Flash variant performs comparably and even secures the highest AUC of 0.8244 on ASSISTments 2009. This implies that performance gains are primarily driven by the quality of retrieved pedagogical contexts and the logic-augmented prompt structure rather than solely by the parametric knowledge of the underlying language model.

\begin{figure*}[!t]
\centering
\includegraphics[width=0.95\linewidth]{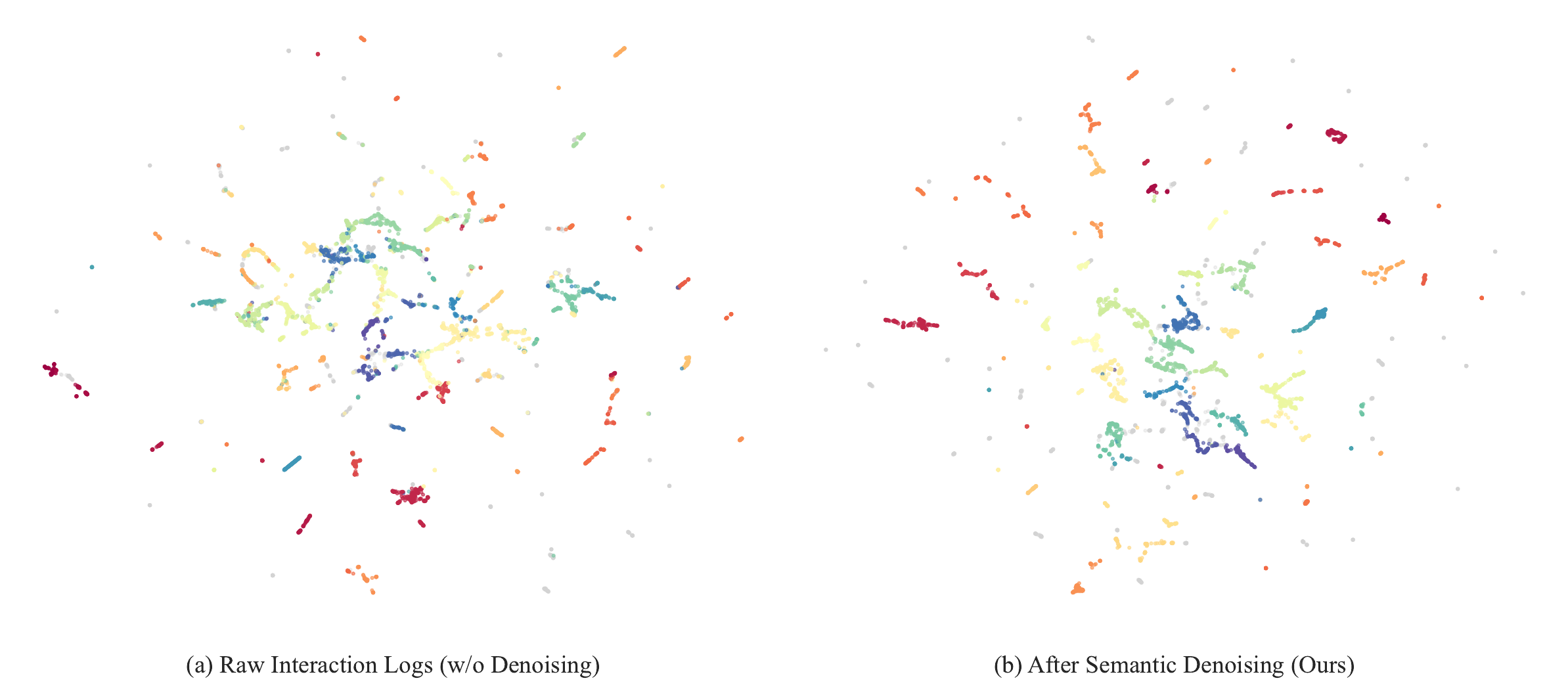}
\caption{UMAP visualization of cognitive schema discovery. (a) Embeddings from raw logs overlap significantly, suggesting statistical artifacts (e.g., IDs, scores) obscure underlying patterns. (b) Conversely, Semantic Denoising yields distinct, compact clusters, confirming effective separation of cognitive signals from noise.}
\label{fig:umap_ablation}
\end{figure*}

\subsection{Ablation Study}

\subsubsection{Decisive Impact of Logic Constraints}
The ablation results unequivocally demonstrate that the Logic-Augmented Inference mechanism is the most critical driver of MERIT's performance. A direct comparison between the full model and the ``w/o Logic Constraints'' variant, which retains retrieval functions but lacks explicit boundary rules, reveals substantial performance gaps. For example, eliminating logic constraints leads to a significant performance decline on the BePKT dataset (GPT-4o backbone), with the AUC plummeting by over 18\%—from 0.8036 to 0.6203. Similarly, on ASSISTments 2009 with Gemini-Flash, the AUC falls from 0.8244 to 0.7015. This pattern suggests that even with high-quality retrieved context, standard LLMs are prone to momentum bias, where they blindly predict success based on recent history. Consequently, the explicit ``Spike Rule'' proves essential for regulating reasoning during difficulty transitions.

\subsubsection{Inefficacy of Naive Retrieval}
The experiments highlight that unstructured retrieval strategies fail to enhance knowledge tracing performance. Notably, the ``w/o Global Routing'' variant utilizing flat retrieval and the ``w/o Interpretative Traces'' variant relying on raw history perform nearly identically to the ``w/o Retrieval'' baseline across most datasets. For example, on ASSISTments 2009 with Gemini-Flash, the baseline achieves an AUC of 0.6708, while removing Global Routing or Interpretative Traces yields AUCs of 0.6715 and 0.6709, respectively. This indicates that simply retrieving raw, unorganized historical data introduces noise rather than signal. The substantial leap to 0.8244 in the full model confirms that the value of the memory bank lies strictly in its structure via Global Routing and its content through Interpretative Traces, which jointly filter noise and provide crystallized reasoning signals.

\subsubsection{Limitations of Parametric Knowledge}
The baseline performance of the ``w/o Retrieval'' setting reveals the inherent ceiling of relying solely on the internal parametric knowledge of LLMs. Without external memory, even powerful models like GPT-4o struggle to capture the nuances of student learning trajectories, yielding an AUC of only 0.6079 on ASSISTments 2009. The consistent and significant improvements observed in the full MERIT framework, which reaches an AUC of 0.8188 with GPT-4o, validate the fundamental hypothesis that domain-specific, retrieved cognitive contexts are indispensable for bridging the gap between general-purpose reasoning and precise educational diagnosis.

\subsection{Parameter Sensitivity Analysis}

\subsubsection{Impact of Retrieval Volume}
To evaluate the trade-off between context sufficiency and information noise, we conduct sensitivity experiments on the ASSISTments 2009 dataset using the Gemini-2.5-Flash backbone. We vary the number of retrieved exercises ($k$) and observe the performance trends. As illustrated in Figure \ref{fig:sensitivity}(a), expanding the retrieval window initially improves diagnostic capability, confirming that multiple analogous exercises provide necessary pedagogical context. However, increasing the sample size beyond $k=3$ leads to performance degradation. This suggests that excessive retrieval introduces irrelevant noise that hinders the reasoning process. Consequently, a compact retrieval set is preferable, ensuring the model focuses on pertinent logic patterns.

\subsubsection{Balancing Semantic and Symbolic Signals}
We further analyze the hybrid search weight $\alpha$, which controls the balance between latent semantic features and explicit symbolic keywords, under the same experimental setting. As shown in Figure \ref{fig:sensitivity}(b), the model achieves its best performance at $\alpha=0.7$. Moving away from this value is associated with a noticeable performance decline. Specifically, assigning a lower weight limits the benefit from structural constraints introduced by symbolic matching, while a higher weight reduces the contribution of semantic relationships captured by vector embeddings. These observations suggest that properly balancing semantic similarity and symbolic constraints is important for robust knowledge tracing.

\subsection{Visualization of Cognitive Schemas}
To qualitatively assess the Cognitive Schema Discovery module, we visualize the latent manifold of student interactions using Uniform Manifold Approximation and Projection (UMAP) \cite{mcinnes2018umap}. We compare embedding distributions generated from raw interaction logs against those processed by the Semantic Denoising mechanism. To ensure consistency, points in both plots are colored according to the cognitive schemas identified in the final denoised space.

Figure \ref{fig:umap_ablation}(a) shows that representations derived from raw logs lack distinct structure, with substantial overlap between different cognitive profiles. This overlap suggests that without regularization, the embedding model overfits high-frequency statistical artifacts, such as system identifiers and raw numerical scores, creating spurious correlations between students with unrelated pedagogical needs. In contrast, Figure \ref{fig:umap_ablation}(b) demonstrates that the semantic denoising pipeline effectively removes these variables. The resulting manifold displays a structured topology where students form compact, well-separated clusters. These clusters correspond to coherent cognitive schemas, such as specific error patterns in Geometry or Algebra, demonstrating that the denoising step is essential for recovering the underlying pedagogical structure of the data.

\section{Conclusion}
\label{sec:conclusion}

We present MERIT, a training-free framework that integrates the predictive strength of deep learning with the interpretability of cognitive modeling. By decoupling reasoning from knowledge storage, the framework helps mitigate context and memory limitations commonly observed in educational LLM applications. Instead of relying on computationally expensive fine-tuning, our approach employs a frozen LLM to reason over a structured and interpretable offline memory bank. This design achieves strong performance across diverse benchmarks while enabling Knowledge Tracing to move beyond black-box probability estimation toward a more transparent, evidence-based diagnostic process.

Furthermore, MERIT is well suited for scalable real-world deployment. Its non-parametric design allows incremental student data to be incorporated while reducing the risk of catastrophic forgetting and avoiding repeated retraining. Experimental results show that retrieving structured reasoning traces yields better performance than both traditional latent vector models and unconstrained LLM prompting. Future work will explore the use of MERIT in white-box intelligent tutoring systems, where models act as transparent pedagogical partners that provide actionable insights to educators and learners.

%%
%% Bibainliography
\bibliographystyle{ACM-Reference-Format}
\bibliography{main}

\appendix

\section{Detailed Prompt Specifications}
\label{sec:appendix_prompts}

This appendix provides the comprehensive prompt templates utilized in the MERIT framework. These templates are crucial for the reproducibility of our experiments and encompass the three core stages of our pipeline:

\begin{itemize}
    \item \textbf{Stage I: Cognitive Schema Discovery.} Guiding LLM to perform semantic denoising on raw interaction logs and identify the student's latent cognitive schema.
    \item \textbf{Stage II: Interpretative Paradigm Construction.} Directing the LLM to generate structured, interpretable "Annotated Cognitive Paradigms" (CoT rationales) from offline training data, which includes ground truth outcomes.
    \item \textbf{Stage III: Logic-Augmented Online Inference.} The final retrieval-augmented generation prompt used during inference, integrating the target student's history, retrieved memory paradigms, and explicit logic constraints (e.g., the Spike Rule).
\end{itemize}

\section{Notation Table}
\label{sec:appendix_notation}

Table \ref{tab:notation} summarizes the key mathematical notations used throughout this paper.

\section{Case Study: Mitigating Momentum Bias via Interpretative Memory}
\label{sec:case_study}

To demonstrate how the Interpretative Memory Bank resolves ambiguity, we analyze a representative case from the ASSISTments 2009 test set (Student ID: 712). Table \ref{tab:case_study_trace} details the execution trace.

The student presents a potentially misleading history: five consecutive correct responses on questions related to \textit{Range}. While this appears to indicate mastery, a closer inspection reveals that all questions share the minimum possible difficulty ($d=0.0$). This creates a classic "illusion of competence." Traditional deep learning models, such as DKT, primarily track response correctness ($r_t=1$) and misinterpret this streak as high proficiency. Consequently, the baseline model predicts success ($P=0.85$) for the subsequent target question, failing to account for the student's inability to handle a sudden concept shift to \textit{Median} at maximum difficulty ($d=1.0$).

MERIT avoids this pitfall by retrieving relevant evidence rather than relying solely on sequence momentum. First, the Cognitive Schema Discovery module maps the student to a "Fragile Mastery" cluster, recognizing that their success is limited to low-difficulty items. Second, the system retrieves a structurally identical historical paradigm (Paradigm \#5 from Cluster \#5). This entry describes a peer who similarly maintained a perfect streak on trivial statistical plots ($d=0.0$) but failed immediately when facing a Venn Diagram task at maximum difficulty ($d=1.0$). By injecting this paradigm into the inference prompt, the reasoning engine draws a direct parallel. The retrieved analysis warns that the trivial questions created a "misleading sense of momentum," allowing MERIT to identify the high risk associated with the difficulty spike.

Supported by this external memory, MERIT correctly predicts the failure, dropping the probability to $0.35$ and aligning with the ground truth. This case underscores that the Interpretative Memory Bank serves as a reference anchor, enabling the model to look beyond immediate history and use collective pedagogical experiences to generate precise, evidence-backed diagnoses.

\begin{onecolumn}

% =================================================================================
% 1. Cognitive Schema Discovery Prompt
% =================================================================================
\begin{tcolorbox}[
    breakable,
    enhanced,
    colback=white,
    colframe=black!80,
    colbacktitle=black!85,
    coltitle=white,
    title=Stage I: Prompt for Cognitive Schema Discovery,
    fonttitle=\large\bfseries,
    left=8pt, right=8pt, top=8pt, bottom=8pt,
    boxrule=0.8pt,
    sharp corners,
    overlay unbroken and first={
        \draw[line width=0.8pt, black!80] (frame.south west) -- (frame.south east);
    },
    overlay middle and last={
        \draw[line width=0.8pt, black!80] (frame.north west) -- (frame.north east);
        \draw[line width=0.8pt, black!80] (frame.south west) -- (frame.south east);
    }
]
\small

You are an expert educational psychologist and data scientist. Your task is to analyze a student's learning history to identify their underlying "Cognitive Schema."

\vspace{6pt}
\noindent\textbf{Input Data}
\hrule height 0.5pt \vspace{3pt}
You will be provided with a raw sequence of a student's interactions.
Each interaction contains:
\begin{itemize}[leftmargin=12pt]
    \item \textbf{Question Content}: The text or key concepts of the problem.
    \item \textbf{Response}: Correct (1) or Incorrect (0).
    \item \textbf{Difficulty}: A normalized difficulty score (0.0 to 1.0).
\end{itemize}

\vspace{6pt}
\noindent\textbf{Analysis Goals}
\hrule height 0.5pt \vspace{3pt}
Your goal is to perform "Semantic Denoising" and "Schema Identification":

1.  \textbf{Semantic Denoising}: Ignore superficial statistical artifacts (e.g., exact timestamps, system IDs, and minor difficulty fluctuations like 0.12 vs 0.13). Focus strictly on the \textit{concepts} and \textit{behavioral patterns} of success/failure.

2.  \textbf{Schema Identification}: Summarize the student's cognitive profile into a coherent schema label.

\vspace{6pt}
\noindent\textbf{Output Format}
\hrule height 0.5pt \vspace{3pt}
Provide a concise summary (no more than 50 words) that serves as a semantic label for this student.
The label should capture:
\begin{itemize}[leftmargin=12pt]
    \item \textbf{Strengths}: Concepts the student has mastered.
    \item \textbf{Weaknesses}: Concepts where the student struggles.
    \item \textbf{Behavioral Pattern}: e.g., "Careless errors on easy tasks", "Struggles with high-difficulty geometry", and "Solid mastery but slow learner".
\end{itemize}

\textbf{Example Input Sequence:}
\texttt{[("Fraction Addition", 1, 0.3), ("Fraction Multiplication", 0, 0.4), ("Geometry Area", 1, 0.8), ("Fraction Division", 0, 0.5)]}

\textbf{Example Output Schema:}
"Proficient in Geometry and basic Arithmetic but exhibits persistent conceptual gaps in intermediate Fraction operations, specifically multiplication and division."

\end{tcolorbox}

\vspace{12pt}

% =================================================================================
% 2. Paradigm Construction Prompt
% =================================================================================
\begin{tcolorbox}[
    breakable,
    enhanced,
    colback=white,
    colframe=black!80,
    colbacktitle=black!85,
    coltitle=white,
    title=Stage II: Prompt for Interpretative Paradigm Construction,
    fonttitle=\large\bfseries,
    left=8pt, right=8pt, top=8pt, bottom=8pt,
    boxrule=0.8pt,
    sharp corners,
    overlay unbroken and first={
        \draw[line width=0.8pt, black!80] (frame.south west) -- (frame.south east);
    },
    overlay middle and last={
        \draw[line width=0.8pt, black!80] (frame.north west) -- (frame.north east);
        \draw[line width=0.8pt, black!80] (frame.south west) -- (frame.south east);
    }
]
\small

You are a pedagogical expert tasked with creating an "Annotated Cognitive Paradigm" for a student's learning trajectory.
This paradigm will be stored in a memory bank to help interpret future students' behaviors.

\vspace{6pt}
\noindent\textbf{Input Context}
\hrule height 0.5pt \vspace{3pt}
\begin{itemize}[leftmargin=12pt]
    \item \textbf{Student History}: A sequence of past interactions (Question, Result, Difficulty).
    \item \textbf{Target Question}: The next question the student attempted.
    \item \textbf{Ground Truth Outcome}: Whether the student \textbf{actually} answered the target question correctly (0 or 1).
\end{itemize}

\vspace{6pt}
\noindent\textbf{Annotation Task}
\vspace{1pt}\hrule height 0.5pt \vspace{3pt}
Generate a structured analysis that explains \textit{why} the student achieved this specific outcome.
Your analysis must include four specific components:

1.  \textbf{Knowledge State ($\mathbf{k}$)}: A summary of the student's current mastery level based on history.

2.  \textbf{Key Pattern ($\mathbf{p}$)}: Identify the behavioral archetype. Choose one from:
    \begin{itemize}
        \item \textit{"Solid Mastery"}: Consistent correctness across difficulties.
        \item \textit{"Difficulty Spike Failure"}: Failed because the jump in difficulty was too high compared to history.
        \item \textit{"Careless Slip"}: Failed an [EASY] item despite knowing harder ones.
        \item \textit{"Concept Gaps"}: Consistent failure in a specific topic.
        \item \textit{"Lucky Guess"}: Correct on a [HARD] item despite weak history.
    \end{itemize}
3.  \textbf{Difficulty Context ($\mathbf{d}$)}: Assess the target question's difficulty relative to the student's history (e.g., "Significantly harder than previous items").

4.  \textbf{Causal Reasoning ($\mathbf{r}$)}: A step-by-step logical chain connecting the history to the outcome.

\vspace{6pt}
\noindent\textbf{Output Format (Strict JSON)}
\hrule height 0.5pt \vspace{3pt}
\begin{verbatim}
{
  "knowledge_state": "Student shows strong grasp of algebra but weak geometry.",
  "key_pattern": "Difficulty Spike Failure",
  "difficulty_context": "Target is [HARD] while history is mostly [EASY].",
  "causal_reasoning": "The student has only practiced basic concepts. The sudden increase in difficulty……to failure.",
  "summary": "Failure due to lack of exposure to high-difficulty items."
}
\end{verbatim}

\end{tcolorbox}

\vspace{12pt}

% =================================================================================
% 3. Online Inference Prompt
% =================================================================================
\begin{tcolorbox}[
    breakable,
    enhanced,
    colback=white,
    colframe=black!80,
    colbacktitle=black!85,
    coltitle=white,
    title=Stage III \& IV: Prompt for Logic-Augmented Online Inference,
    fonttitle=\large\bfseries,
    left=8pt, right=8pt, top=8pt, bottom=8pt,
    boxrule=0.8pt,
    sharp corners,
    overlay unbroken and first={
        \draw[line width=0.8pt, black!80] (frame.south west) -- (frame.south east);
    },
    overlay middle and last={
        \draw[line width=0.8pt, black!80] (frame.north west) -- (frame.north east);
        \draw[line width=0.8pt, black!80] (frame.south west) -- (frame.south east);
    }
]
\small

You are an advanced Knowledge Tracing system (MERIT). Your goal is to predict whether a target student will answer the next question correctly.

\vspace{6pt}
\noindent\textbf{Student Context}
\vspace{1pt}\hrule height 0.5pt \vspace{3pt}
\textbf{Target Student History:} \\
\texttt{\{student\_history\_sequence\}}

\textbf{Next Question (Target):} \\
\texttt{\{target\_question\_content\}} \\
\textbf{Difficulty Level:} \texttt{\{difficulty\_tag\}} (Tags: [EASY], [MEDIUM], [HARD])

\vspace{6pt}
\noindent\textbf{Retrieved Interpretative Memory (from Paradigm Bank)}
\hrule height 0.5pt \vspace{3pt}
The following examples are retrieved from students with similar \textbf{Cognitive Schemas} to the target student. Use them as reference points for your reasoning to reduce hallucinations.

\textbf{--- Paradigm 1 ---} \\
\textbf{History Summary}: \texttt{\{retrieved\_history\_1\}} \\
\textbf{Pattern}: \texttt{\{retrieved\_pattern\_1\}} \\
\textbf{Outcome}: \texttt{\{retrieved\_outcome\_1\}} \\
\textbf{Reasoning}: \textbf{\texttt{\{retrieved\_reasoning\_1\}}}

\textbf{--- Paradigm 2 ---} \\
...

\vspace{6pt}
\noindent\textbf{Logic Constraints (The Spike Rule)}
\hrule height 0.5pt \vspace{3pt}
You must strictly adhere to the following logic to prevent "Momentum Bias":
\begin{itemize}[leftmargin=12pt]
    \item \textbf{IF} the student has a recent streak of correct answers on [EASY] items,
    \item \textbf{AND} the target question is labeled [HARD],
    \item \textbf{THEN} do NOT blindly predict "Correct" based on momentum. You must consider the "Difficulty Spike" pattern. There is a high probability of failure unless there is specific evidence of advanced mastery.
\end{itemize}

\vspace{6pt}
\noindent\textbf{Prediction Task}
\hrule height 0.5pt \vspace{3pt}
Based on the student's history, the retrieved paradigms, and the logic constraints:
1.  Analyze the student's current knowledge state.
2.  Compare the target scenario with the retrieved paradigms.
3.  Apply the Logic Constraints (Spike Rule).
4.  Predict the outcome (0 for Incorrect, 1 for Correct) and provide the probability.

\vspace{6pt}
\noindent\textbf{Output Format (Strict JSON)}
\hrule height 0.5pt \vspace{3pt}
\begin{verbatim}
{
  "reasoning_trace": "Step-by-step analysis incorporating memory and logic...",
  "prediction": 0,
  "probability": 0.xx
}
\end{verbatim}

\end{tcolorbox}
\end{onecolumn}

\twocolumn

% 使用 table* (带星号) 让表格横跨双栏，置于页面顶部 [t]
\begin{table*}[t]
\centering
\caption{Summary of key notations.}
\label{tab:notation}
\renewcommand{\arraystretch}{1.2} % 稍微增加行高，防止数学符号拥挤
% 使用 tabularx 占满整个文本宽度 (\textwidth)
% 'l' 列用于符号 (左对齐)，'X' 列用于描述 (自动换行，防止出页)
\begin{tabularx}{\textwidth}{l|X}
\toprule
\textbf{Symbol} & \textbf{Description} \\
\midrule
\multicolumn{2}{l}{\textit{\textbf{Problem Definition}}} \\
$\mathcal{U}, \mathcal{E}$ & Set of students and set of exercises, respectively. \\
$\mathcal{S}_u$ & Interaction sequence for student $u$: $\mathcal{S}_u = \{s_1, s_2, \dots, s_t\}$. \\
$s_t = (e_t, r_t, d_t)$ & Interaction tuple at step $t$, consisting of exercise content $e_t$, response $r_t$, and difficulty $d_t$. \\
$r_t \in \{0, 1\}$ & Binary response correctness (0 indicates Incorrect, 1 indicates Correct). \\
$d_t \in [0, 1]$ & Continuous difficulty scalar for exercise $e_t$. \\
$\hat{y}_{t+1}$ & Predicted probability of correct response for the next exercise. \\
\midrule
\multicolumn{2}{l}{\textit{\textbf{Stage 1: Cognitive Schema Discovery}}} \\
$\tilde{T}_u$ & Denoised text sequence after applying token regularization (Semantic Denoising). \\
$\mathbf{h}_u \in \mathbb{R}^d$ & High-dimensional semantic embedding vector for student $u$. \\
$\mathcal{C} = \{C_1, \dots, C_K\}$ & Set of $K$ latent cognitive clusters identified via density-based clustering. \\
$\mathbf{\mu}_k$ & The centroid vector of cognitive cluster $C_k$. \\
\midrule
\multicolumn{2}{l}{\textit{\textbf{Stage 2 \& 3: Memory Construction \& Retrieval}}} \\
$\mathcal{B}$ & The Interpretative Memory Bank containing paradigm entries $\{\mathcal{M}_1, \dots, \mathcal{M}_N\}$. \\
$\mathcal{M} = \langle \mathcal{S}_u, y, \mathcal{A} \rangle$ & A single memory entry consisting of history, outcome, and generated annotation. \\
$\mathcal{A} = (\mathbf{k}, \mathbf{p}, \mathbf{d}, \mathbf{r})$ & Annotation tuple: Knowledge State, Key Pattern, Difficulty Context, and Causal Reasoning. \\
$\alpha$ & Hybrid search weight ($0 \le \alpha \le 1$) balancing semantic similarity and symbolic keyword matching. \\
$\mathcal{R}_u$ & Set of top-$N$ retrieved memory entries relevant to the target student. \\
\midrule
\multicolumn{2}{l}{\textit{\textbf{Stage 4: Logic-Augmented Inference}}} \\
$L(d_t)$ & Discrete difficulty function mapping scalar scores to tags (e.g., \texttt{[HARD]}). \\
$\tau$ & Similarity threshold used for filtering irrelevant retrieved contexts. \\
\bottomrule
\end{tabularx}
\end{table*}

\begin{table*}[t]
\centering
\caption{Detailed execution trace for the case study (Student ID: 712). The table contrasts the surface-level observation that misled the baseline model with the deep, memory-augmented reasoning process of MERIT. By retrieving a structurally similar paradigm (Paradigm \#5), MERIT identifies the "Difficulty Spike" pattern and corrects the prediction.}
\label{tab:case_study_trace}
\renewcommand{\arraystretch}{1.3}
\begin{tabularx}{\textwidth}{l|X}
\toprule
\textbf{Pipeline Stage} & \textbf{System Execution \& Analysis} \\
\midrule
\multicolumn{2}{l}{\cellcolor{gray!10}\textit{Phase 1: Input Analysis (The Trap)}} \\
\textbf{Target Student History} & Sequence: \texttt{[("Range", 0.0, Correct)]} $\times$ 5 \newline
\textit{Observed Pattern:} \textcolor{teal}{Perfect Correct Streak} on minimal difficulty items. \\
\textbf{Target Task} & Concept: \textbf{Median} \quad Difficulty: \textbf{1.0 (Max)} \quad \textit{Risk:} \textcolor{red}{Extreme Difficulty Spike}. \\
\textbf{Baseline Error} & DKT predicts \textbf{High Probability (0.85)} due to strong positive momentum bias. \\
\midrule
\multicolumn{2}{l}{\cellcolor{gray!10}\textit{Phase 2: MERIT Evidence Retrieval (The Reality Check)}} \\
\textbf{Retrieved Paradigm} & \textbf{Source:} Memory Bank Paradigm \#5 (Cluster \#5) \newline
\textbf{Historical Context:} \textit{"Student had perfect streak on trivial items: Stem/Leaf Plot ($d=0.0$), Median ($d=0.0$)..."} \newline
\textbf{Outcome:} \textcolor{red}{False} on next question (Venn Diagram, $d=1.0$). \newline
\textbf{Retrieved Analysis:} \newline
$\hookrightarrow$ \textbf{Key Pattern:} \texttt{"Difficulty Spike Failure"} \newline
$\hookrightarrow$ \textbf{Reasoning:} \textit{"The student's perfect streak on trivial (0.0 difficulty) questions created a \textbf{misleading sense of momentum}. The failure was caused by an extreme and simultaneous spike in concept difficulty (0.0 $\to$ 1.0)."} \\
\midrule
\multicolumn{2}{l}{\cellcolor{gray!10}\textit{Phase 3: Logic-Augmented Inference (The Decision)}} \\
\textbf{Contextual Reasoning} & \textit{LLM CoT Trace:} "The current student matches the retrieved paradigm perfectly. Both show mastery only on trivial tasks ($d=0.0$). The retrieved analysis warns that this momentum is 'misleading' when facing a difficulty spike to 1.0. Therefore, success is unlikely." \\
\textbf{Final Prediction} & \textbf{Prediction:} \textcolor{red}{Incorrect} ($P=0.35$). \quad \textbf{Ground Truth:} \textcolor{red}{Incorrect}. \\
\bottomrule
\end{tabularx}
\end{table*}

\end{document}